\documentclass[conference]{IEEEtran}
\IEEEoverridecommandlockouts
\usepackage{cite}
\usepackage{amsmath,amssymb,amsfonts}
\usepackage{algorithmic}
\usepackage{graphicx}
\usepackage{textcomp}
\usepackage{xcolor}
\renewcommand\textcolor[2]{#2}
\def\BibTeX{{\rm B\kern-.05em{\sc i\kern-.025em b}\kern-.08em
    T\kern-.1667em\lower.7ex\hbox{E}\kern-.125emX}}
    
\usepackage{booktabs}  
\usepackage{multirow,makecell}
\usepackage{subfigure}
\usepackage[ruled,vlined,linesnumbered]{algorithm2e}

\usepackage{etoolbox}
\makeatletter
\patchcmd{\@makecaption}
  {\scshape}
  {}
  {}
  {}
\makeatletter
\patchcmd{\@makecaption}
  {\\}
  {.\ }
  {}
  {}
\makeatother

\newcommand{\M}{Auto-ViT-Acc}

\begin{document}

\title{\LARGE{\M: An FPGA-Aware Automatic Acceleration Framework for Vision Transformer with Mixed-Scheme Quantization}}

\author{%
Mengshu Sun\affmark[1]*, Zhengang Li\affmark[1]*, Alec Lu\affmark[2]*,
Yanyu Li\affmark[1], Sung-En Chang\affmark[1], Xiaolong Ma\affmark[1] \\
Xue Lin\affmark[1], Zhenman Fang\affmark[2]
\affiliation{%
  \institution{%
  \affaddr{\affmark[1]Department of Electrical \& Computer Engineering, Northeastern University, Boston, MA, United States} \\
  \affaddr{\affmark[2]School of Engineering Science, Simon Fraser University, Burnaby, BC, Canada}} \\
  \affaddr{\affmark[1]\{sun.meng, li.zhen, li.yanyu, chang.sun, ma.xiaol, xue.lin\}@notheastern.edu} \\
  \affaddr{\affmark[2]\{alec\_lu, zhenman\}@sfu.ca}
}}

\author{
Zhengang Li\text{$^\dagger$}\textsuperscript{1}, Mengshu Sun\text{$^\dagger$}\textsuperscript{1}\thanks{$^\dagger$These authors contributed equally.},
Alec Lu\textsuperscript{2}, Haoyu Ma\textsuperscript{3}, Geng Yuan\textsuperscript{1}, Yanyue Xie\textsuperscript{1}, Hao Tang\textsuperscript{4}, Yanyu Li\textsuperscript{1},\\
Miriam Leeser\textsuperscript{1}, Zhangyang Wang\textsuperscript{5}, Xue Lin\textsuperscript{1}, Zhenman Fang\textsuperscript{2}\\
\textsuperscript{1}Northeastern University, 
\textsuperscript{2}Simon Fraser University, 
\textsuperscript{3}University of California, Irvine, \\
\textsuperscript{4}ETH Zurich, 
\textsuperscript{5}The University of Texas at Austin \\
{E-mail:} \textsuperscript{1}\{li.zhen, sun.meng, yuan.geng, xie.yany, li.yanyu, xue.lin\}@northeastern.edu, \textsuperscript{1}mel@coe.neu.edu,\\
\textsuperscript{2}\{alec\_lu, zhenman\}@sfu.ca, \textsuperscript{3}haoyum3@uci.edu, \textsuperscript{4}hao.tang@vision.ee.ethz.ch, 
\textsuperscript{5}atlaswang@utexas.edu}

\maketitle

\begin{abstract}


Vision transformers (ViTs) are emerging with significantly improved accuracy in computer vision tasks. However, their complex architecture and enormous computation/storage demand impose urgent needs for new hardware accelerator design methodology. This work proposes an FPGA-aware automatic ViT acceleration framework based on the proposed mixed-scheme quantization. To the best of our knowledge, this is the first FPGA-based ViT acceleration framework exploring model quantization.
Compared with state-of-the-art ViT quantization work (algorithmic approach only without hardware acceleration), our quantization achieves 0.47\% to 1.36\% higher Top-1 accuracy under the same bit-width. Compared with the 32-bit floating-point baseline FPGA accelerator, our accelerator achieves around 5.6$\times$ improvement on the frame rate (i.e., 56.8 FPS vs. 10.0 FPS) with 0.71\% accuracy drop on ImageNet dataset for DeiT-base.

\end{abstract}

\section{Introduction}


Transformer, an attention-based encoder-decoder architecture \cite{vaswani2017attention}, has revolutionized the field of natural language processing (NLP) in the past five years. Inspired by NLP successes, researchers began to adopt transformer-like architecture to computer vision tasks i.e., vision transformers (ViTs), achieving better performance compared with state-of-the-art convolutional neural networks (CNNs) \cite{dosovitskiy2020image, Touvron2021TrainingDI, yuan2021tokens}. However, the complex model architecture and enormous computation and storage of ViT make it a challenging task for their deployment into resource constrained edge devices.


Model quantization, as a crucial technique for DNN interence acceleration on edge devices, has been broadly explored for CNNs~\cite{courbariaux2015binaryconnect, rastegari2016xnor,zhou2016dorefa, choi2018pact,leng2018extremely,li2020additive} with different bit-widths and also different quantization schemes, e.g., fixed-point and power-of-two (PoT). These two types of schemes were mixed in~\cite{chang2021mix} for FPGA-based implementations to fully utilize the hardware computation resources. As for quantization of transformer models, few efforts~\cite{liu2021post} have been devoted to ViTs, while majority of work~\cite{zafrir2019q8bert,zhang2020ternarybert,bai2020binarybert} was still on transformers for NLP with purely algorithmic approaches. There are two open problems for ViT quantization: 1. Do existing quantization schemes for CNNs work well for ViTs? 2. How to systematically determine the bit-width and mixing ratio in mixed-scheme quantization for better accuracy and throughput performance for ViTs?

In this paper, we first explore the feasibility of the well-studied CNN quantization schemes---including fixed-point, PoT, and their mix---on ViT and make the following observations.
First, fixed quantization possesses superior accuracy performance, and its computation can be efficiently implemented with the DSP resources on FPGA.
Second, the PoT scheme offers a highly efficient quantization with still acceptable accuracy, where multiplications can be replaced by simple shift operations, and thus suitable for implementation with LUT resource on FPGA.
Finally, combining fixed-point and PoT has the potential to further improve FPGA resource utilization for inference acceleration while maintaining accuracy.

Based on the above, we develop an FPGA-aware automatic ViT acceleration (\M) framework for our mixed-scheme ViT quantization algorithm.
It contains an ``FPGA Resource Utilization Modeling'' module to give performance analysis and estimate the frame rate (FPS) for the FPGA ViT accelerator under a certain setting of model bit-widths, which will be reduced until the target FPS is achieved.
In this way, the bit-width and the ratio of fixed-point quantized rows over PoT quantized rows can be optimized and used as inputs to guide the quantization algorithm.
This framework also designs a novel FPGA compute engine for ViT multi-head attention with optimizations for accelerators.
We automate the entire workflow based on a target FPS, to obtain a quantized model and an FPGA accelerator.
The contributions of our work are summarized as follows:

\begin{itemize}
    \item \textbf{An FPGA-aware mixed-scheme ViT quantization algorithm that can fully leverage heterogeneous FPGA resources while maximally retaining accuracy.}
    \item \textbf{An automated ViT acceleration framework with FPGA resource utilization modeling to automatically find the best combination of quantization bit-widths and the scheme mixing ratio for a target FPS.}

    \item \textbf{A novel FPGA computing engine for ViT multi-head attention and related accelerator optimizations.}
    \item \textbf{To the best of our knowledge, \M~is the first for ViT acceleration on FPGAs exploring model quantization with significant performance improvements.}
\end{itemize}
\section{Related Work}

\subsection{Vision Transformer}

The ViT architecture was first proposed in \cite{dosovitskiy2020image}, which adopts the self-attention mechanism \cite{vaswani2017attention} for image classification tasks. 
Different from CNNs, ViT interprets an image as a sequence of patches and then inputs to standard transformer encoders as used in NLP. 
However, it requires pre-training with complex and massive datasets such as ImageNet-21k and JFT-300M. 
To address this, DeiT \cite{Touvron2021TrainingDI} and T2T-ViT \cite{yuan2021tokens} were proposed to reduce 
dependency on massive pre-training and achieve better accuracy than ResNets \cite{he2016deep} of comparable size on ImageNet. 




\begin{figure}[htb]
  \centering
  \includegraphics[width=0.9\linewidth]{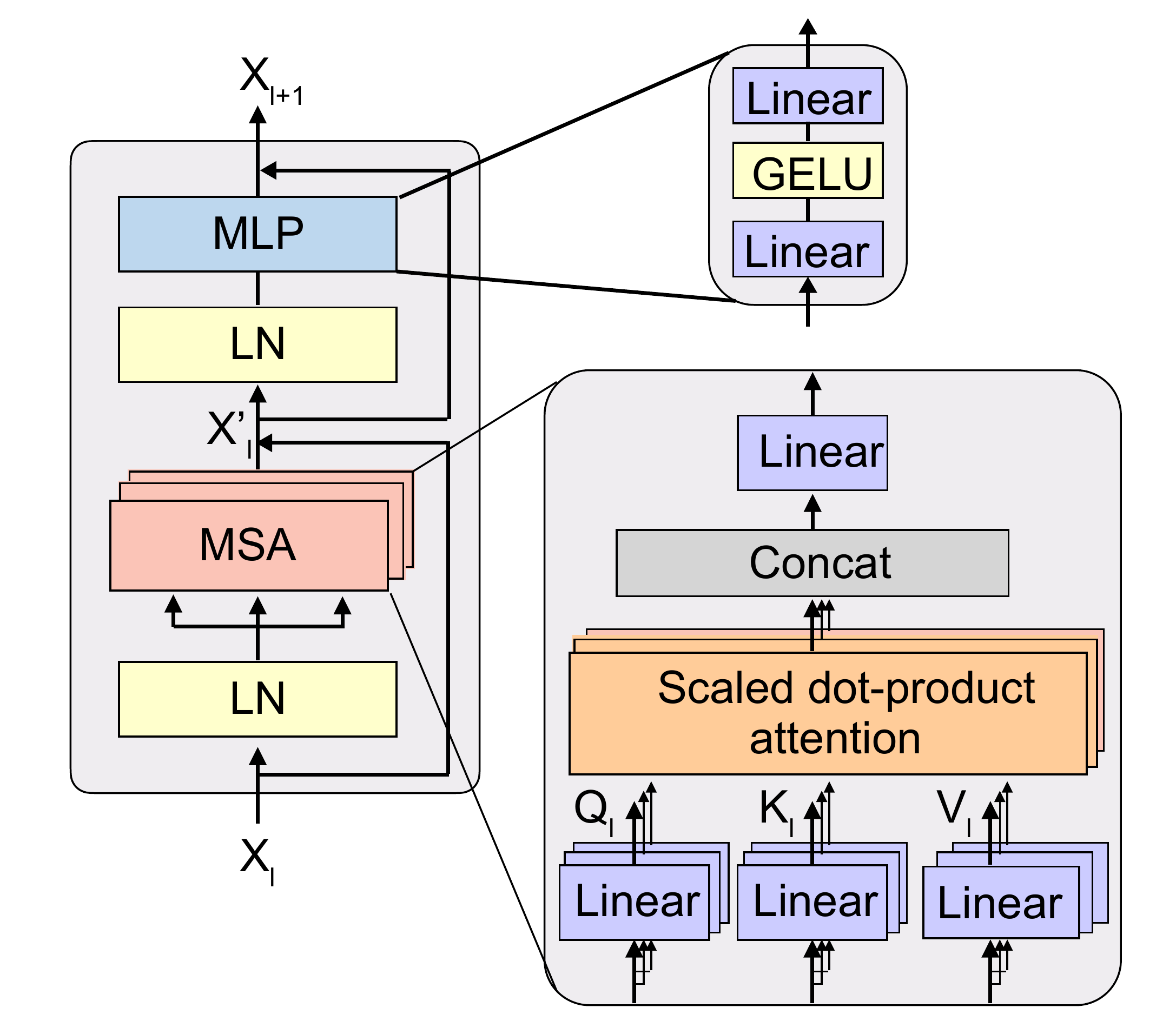}
  \caption{Transformer encoder block structure.}
  \label{fig:encoder}
\end{figure}

In ViT, the main model architecture is transformer encoder blocks with multi-headed self-attention (MSA) and multi-layer perceptron (MLP) blocks as shown in Fig.~\ref{fig:encoder}. The layernorm (LN) is applied prior to MSA and MLP. The encoder block operations are:
\begin{equation}
\begin{aligned}
     \mathbf{X'}_l & =  \text{MSA} ( \text{LN} ( \mathbf{X}_{l} ) ) +  \mathbf{X}_{l}, \\
     \mathbf{X}_{l+1} & =  \text{MLP} ( \text{LN} (\mathbf{X'}_l)) + \mathbf{X'}_l, \\
\end{aligned}
\end{equation}
where $\mathbf{X}_l$ denotes the input sequence of the $l$-th encoder block.


These modules involve large matrix multiplications incurring the most computational cost. Therefore, we quantize all linear layers involved in matrix multiplication, but not layer normalization, due to their low computational cost and potential effects on accuracy. 

\subsection{DNN Model Quantization}

\subsubsection{Quantization Schemes}

Model quantization has been intensively explored for deep neural networks (DNNs) such as CNNs and recurrent neural networks (RNNs).

There are schemes using uniform quantization intervals including binary \cite{courbariaux2015binaryconnect, rastegari2016xnor}
ternary \cite{he2019simultaneously}, 
and low-bit-width fixed-point \cite{zhou2016dorefa, choi2018pact}.
Although binary and ternary quantization significantly reduce operations and simplify hardware implementation, it introduces large accuracy loss. 
The fixed-point quantization scheme, on the other hand, applies modest and flexible quantization rates to preserve accuracy close to that of 32-bit floating-point models.
For instance, 4-bit fixed-point introduces zero or negligible accuracy loss.
Fixed-point quantization scheme was implemented with different methods and algorithms, such as DoReFa-Net \cite{zhou2016dorefa} and PACT \cite{choi2018pact}. 

There are also schemes using non-uniform quantization intervals such as power-of-two (PoT) \cite{leng2018extremely} and additive PoT \cite{li2020additive}, by which multiplications can be replaced with bit shifting operations.
Furthermore, 
PoT presents higher precision around the mean, and therefore better fits the Gaussian distribution of DNN weights \cite{blundell2015weight}.
But it exhibits rigid resolution issue that results in moderate accuracy loss, which cannot be mitigated even with higher bit-width. 
To overcome it, additive PoT was proposed by using a sum of multiple PoT numbers.

\subsubsection{Transformer Quantization}
Quantization has also been applied to transformers, in particular, bidirectional encoder representations from transformers (BERTs) \cite{vaswani2017attention}.
Specifically,~\cite{zafrir2019q8bert} finetuned BERT through 8-bit quantization-aware training. 
The later TernaryBERT~\cite{zhang2020ternarybert} proposed to use an approximation-based and loss-aware ternarization for BERT, and distillation to further reduce accuracy drop caused by lower capacity.
BinaryBERT~\cite{bai2020binarybert} suggested that it is difficult to train a binary BERT directly due to its complex loss landscape and proposed a ternary weight splitting strategy to derive binary BERT with performance as the ternary one. 
All the aforementioned work targeted BERT in NLP tasks, not covering ViT in computer vision tasks. 

A recent work~\cite{liu2021post} evaluated the post-training quantization on ViT and achieved comparable accuracy  as the full-precision version. However, they only used a low quantization rate i.e., 4$\times$, which is equivalent to 8-bit quantization precision. Further, it is a pure algorithmic method and not suitable for acceleration on hardware like FPGAs. 

\subsection{Transformer Accelerators on FPGAs}\label{sec:transformerFPGA}

Recently, weight pruning approaches have also been applied for transformer acceleration on FPGAs.
The study in \cite{li2020ftrans} leveraged block-circulant matrix-based weight representation and FFT/IFFT-based processing elements for matrix-vector multiplication for fully-connected (FC) layers. 
Block-based weight pruning was applied in \cite{qi2021accommodating} to accelerate transformers on FPGAs. 
\cite{zhang2021algorithm} proposed a structural pruning method with memory footprint awareness to compress weights to similar sizes. 
This method effectively compresses the attention mechanism and achieves efficient deployment of data buffers and computing kernels on FPGAs.
Differently, our work explores model quantization for ViT acceleration on FPGA and is orthogonal and complementary to pruning-based prior arts.
\section{New Challenges and Novelty}

ViTs leverage the attention mechanism~\cite{vaswani2017attention} to fulfill various computer vision tasks. 
Compared to CNNs that operate on a fixed-size window with restricted spatial interactions, ViT allows data at all the positions in an image to interact through transformer encoder blocks and thus improving accuracy~\cite{raghu2021vision}.
As mentioned in~\cite{Touvron2021TrainingDI}, ViTs can perform better than representative CNNs like ResNet~\cite{he2016deep} and ResNeXt~\cite{xie2017aggregated}. For instance, DeiT-small with a comparable number of parameters and operations as ResNet-50 achieves higher accuracy than ResNeXt-101, whose size is around 4$\times$ as that of ResNet-50. DeiT-base with comparable size as ResNeXt-101 achieves much higher accuracy.






Although with significant accuracy improvement, there exist challenges in hardware acceleration of ViTs, especially on resource-limited edge devices.
First, even the light-weight DeiT-small model is already a large model for edge devices.
Furthermore, the complexity of multi-head self-attention brings in a new optimization dimension of hardware parallelism. (Detailed discussions are provided in Sec.~\ref{sec:compute_engine}.)
Therefore, model compression techniques including pruning and quantization become essential in ViT hardware acceleration.
Unlike most prior arts mentioned in Sec.~\ref{sec:transformerFPGA}, this paper focuses on model quantization for ViT hardware acceleration. 

Existing work on ViT quantization~\cite{liu2021post} adopted the fixed-point quantization scheme with 8-bit precision. In this paper, we observe that leveraging PoT quantization, which allows multiplications to be replaced by simple shift operations, can achieve better inference performance on FPGAs with the LUT resources, with negligible accuracy loss. Moroever, when combining the fixed-point quantization that mainly consumes the DSP resources on FPGAs, there is more potential to fully utilize the FPGA resource for even better performance.

Besides various quantization schemes, layer-wise multi-precision quantization has been well investigated in \cite{wang2019haq,uhlich2019mixed,dong2019hawq} that assign precisions onto weights and activations of individual layers.
However, as pointed out in~\cite{chang2021mix}, this type of quantization is incompatible with layer-by-layer inference execution on hardware accelerators since it introduces non-uniformality among layers.
In contrast, this paper adopts the \emph{mixed-scheme quantization within each layer}, with a mixture of fixed-point and PoT schemes.
Different from~\cite{chang2021mix} that focuses on CNN acceleration, we use PoT in replacement of their Sum-of-PoT for improved computation efficiency while avoiding compromising accuracy.
Unlike \cite{wang2019haq,uhlich2019mixed,dong2019hawq} which deal with a large search space for precision assignment, we propose a practical mixed-scheme ViT quantization algorithm that closely coordinates with the FPGA-based accelerator design. 

For mixed-scheme quantization, we need the co-design of the quantization algorithm and the FPGA accelerator. We propose a set of automated mechanism (Sec.~\ref{sec:overall_flow}) with FPGA resource utilization modeling to automatically find the best combination of quantization bit-widths and mixed-scheme ratio for a targeted FPS.
Furthermore, from the hardware design aspect, we have the following observations:
First, to prevent extra hardware overhead on output shifting among two schemes i.e., fixed-point and PoT, we propose to align the outputs from two quantization schemes by deriving the relation between their precisions i.e., bit-widths.
Explanations are in Sec. \ref{sec:overall_flow}. 
Second, we propose to use the same ratio of fixed-point to PoT for each head of the MSA module to fully exploit parallelism of FPGA.


\section{FPGA-Aware Mixed-Scheme ViT Quantization Algorithm}

\subsection{Quantization Scheme and Precision}

We propose to use a mixture of fixed-point and PoT within each layer.
Note that we apply quantization only to linear layers of ViT, which involve the most computation-intensive matrix multiplications. We do not quantize for softmax and layer normalization, due to their low computational cost.
Fixed-point quantization scheme has superior accuracy performance, and its computation can be implemented efficiently with DSP resources on FPGA.
PoT is a highly efficient quantization scheme with still acceptable accuracy, where multiplications can be replaced by bit shifting operations, and thus suitable for implementation with LUT resource on FPGA.
Combining fixed-point and PoT can increase FPGA resource utilization to speed up inference, at the same time, retain accuracy.

We use $\mathcal{\prod}^\text{Fixed}_{(b, \alpha)}$ and $\mathcal{\prod}^\text{PoT}_{(b, \alpha)}$ to represent the fixed-point and PoT quantizers, respectively, where $b$ denotes the bit-width and $\alpha$ denotes scaling factor. Detailed quantizer functions can be found in \cite{chang2021mix}. 
In general, a quantizer function maps a floating-point value into a fixed-point or PoT quantized value, equal to multiplication of the scaling factor with a quantization level represented by a $b$-bit number.
For both quantization schemes, $b$-bit number representation corresponds to $2^b-1$ quantization levels (with 1-bit for sign).
{As for the selection of precision or bit-width, to avoid the large search space of scheme and precision assignment and to preserve hardware uniformity among layers, we specify the precision candidates as:  $b$-bit for fixed-point quantized weights, $b'$-bit for PoT quantized weights, and $b$-bit for activations.\footnote{Even for PoT scheme, only weights are PoT quantized and corresponding activations are still fixed-point quantized in order to replace multiplication with bit shifting.}}

\begin{algorithm}[htb]
\caption{FPGA-aware mixed-scheme ViT quantization.} \label{algo: quantization}
\small
\SetKwInOut{Input}{input}
\SetKwInOut{Output}{output}
\SetKwFunction{SGD}{SGD}

\Input{32-bit floating-point pre-trained ViT model $\mathcal{M}$ with weights $\mathbf{W}$; bit-width for fixed-point $b$; bit-width for PoT $b^\prime$; ratio of PoT quantized rows in each layer $k_\mathrm{PoT}$;
}
\Output{Quantized model $\hat{\mathcal{M}}$. }
\ForEach{batch}{
\tcp{forward propagation}
\ForEach{layer $i$ \textnormal{in} $\mathcal{M}$}{
\tcp{calculate variance for each row}
\ForEach{\textnormal{row} $\mathbf{W}_{ij}$ \textnormal{in} layer $i$}{
$\text{var}_{ij} \leftarrow variance(\mathbf{W}_{ij})$ ;
}
\tcp{assign weight quantization scheme for rows} 
\ForEach{\textnormal{row} $\mathbf{W}_{ij}$ in layer $i$}{
\eIf{$\text{var}_{ij}$ belongs to the bottom $k_{\mathrm{PoT}}$ group}
{
$\hat{\mathbf{W}}_{ij} \leftarrow \mathcal{\prod}^\text{PoT}_{b^\prime}(\mathbf{W}_{ij})$;
}
{
$\hat{\mathbf{W}}_{ij} \leftarrow \mathcal{\prod}^\text{Fixed}_{b}(\mathbf{W}_{ij})$;
}
}
$\mathbf{A}_i \leftarrow  \hat{\mathbf{W}}_{i}\cdot \hat{\mathbf{A}}_{i-1}$;\\
\tcp{quantize activations}
$\hat{\mathbf{A}}_i\leftarrow \mathcal{\prod}^\text{Fixed}_{b}(\mathbf{A}_i)$; \\
}
\tcp{backward propagation}
\ForEach{layer $i$ (reverse order)}{
$\frac{\partial Loss}{\partial \mathbf{W}_i} \leftarrow \frac{\partial Loss}{\hat{\mathbf{W}}_{i}}$; \\
$\frac{\partial Loss}{\partial input_i} \leftarrow \frac{\partial Loss}{\hat{\mathbf{A}}_{i-1}}$; 
}} 
Return $\hat{\mathcal{M}}\leftarrow\mathcal{M}\{\hat{\mathbf{W}}\}$.
\end{algorithm}

\subsection{Proposed ViT Quantization Algorithm}  \label{sec:mixed_scheme}






As shown in Algorithm~\ref{algo: quantization}, 
our proposed FPGA-aware mixed-scheme ViT quantization algorithm performs quantization training with given bit-widths i.e., $b$ and $b'$, and the ratio of PoT quantized rows i.e., $k_{pot}$ in each layer (the rest rows are fixed-point quantized).
We use the same ratio $k_{pot}$ among different heads of the MSA module to fully exploit the parallelism of FPGA.
$b$, $b'$, and $k_{pot}$ are determined from our \M~framework.
The quantization scheme is assigned down to the row level of a weight matrix based on the weight distribution. 
In general, if a row has a smaller variance, the PoT scheme is assigned; and otherwise, the fixed-point scheme is assigned.
\section{Proposed \M~Framework} \label{sec:framework}


This section first gives an overview of \M, and then discusses the optimization techniques in the ViT computation engine (Sec.~\ref{sec:compute_engine} and~\ref{sec:other_layers}), and finally provides FPGA resource modeling to determine $b$, $b'$, and $k_{\mathrm{pot}}$ for target frame rate (FPS) 
(Sec.~\ref{sec:accl_framework}).

\subsection{Overview and Design Space Exploration} \label{sec:overall_flow}

\begin{figure*}[htb]
  \centering
  \includegraphics[width=0.8\linewidth]{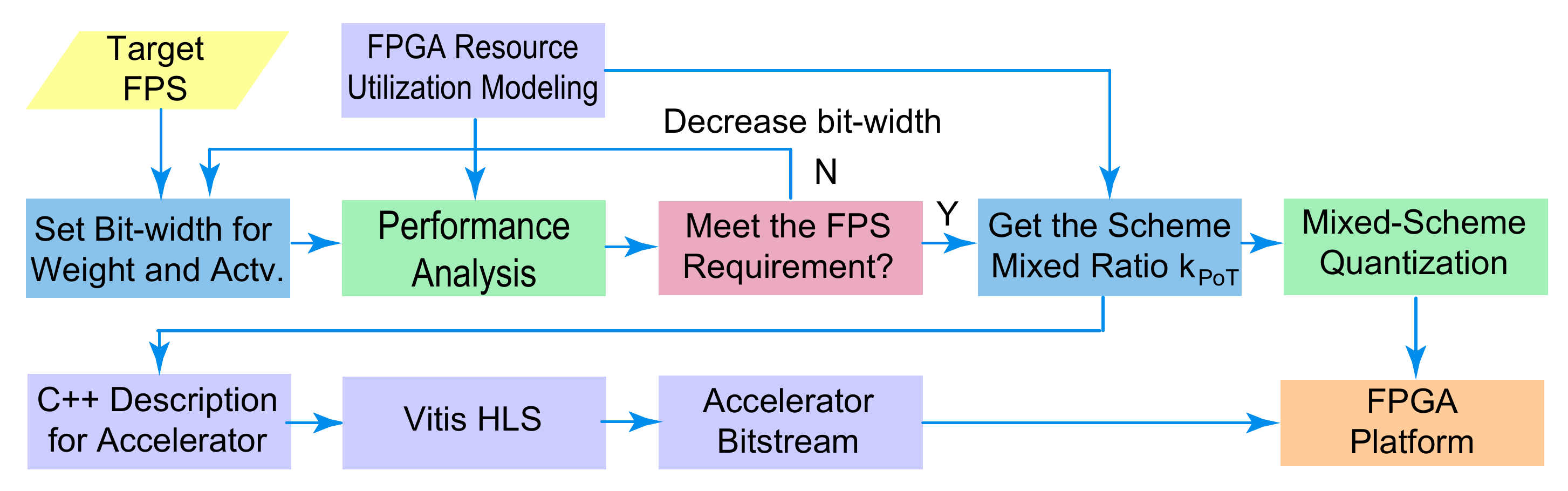}
  \caption{Overview of \M~framework.}
  \label{fig:auto}
\end{figure*}

Fig.~\ref{fig:auto} provides the workflow of our \M~framework for automatic generations of ViT accelerators. 
We start from ``FPGA Resource Utilization Modeling'' module to give performance analysis and estimate the frame rate (FPS) of FPGA ViT accelerator with given bit-widths for the Fixed and PoT schemes i.e., $b$ and $b'$.
We reduce the bit-widths until fulfilling the target FPS. 
The details of resource modeling and performance analysis are discussed in Section~\ref{sec:accl_framework}, which also derive the desired ratio for PoT quantized rows $k_{\mathrm{pot}}$. 
Then our proposed mixed-scheme ViT quantization algorithm uses $b$, $b^\prime$, and $k_{\mathrm{pot}}$ to derive quantized ViT model, which will be implemented on FPGA by going through ``C++ Description for Accelerator'', ``Xilinx Vitis High-Level Synthesis (HLS)'', and ``Accelerator Bitstream''.

About bit-widths, for each layer, we quantize some of the rows into Fixed with $b$-bit for weights and $b$-bit for the corresponding activations i.e., Fixed W[$b$]A[$b$], and quantize the rest rows into PoT  W[$b^\prime$]A[$b$] i.e., $b^\prime$-bit for weights and $b$-bit for corresponding activations.
To prevent extra hardware overhead on output shifting among two schemes, we propose to align the outputs from two schemes by setting $2^{(b^\prime-1)} \leq b$, i.e., if $b$-bit is used for Fixed, then $b^\prime = \lfloor \log_2 b \rfloor + 1$ is used for PoT. For example, the 4-bit Fixed scheme matches the 3-bit PoT scheme, and 8-bit Fixed scheme matches the 4-bit PoT scheme.
This is because the product of the Fixed W[$b$]A[$b$] multiplication has a bit-width of $2 \cdot b$, and the same output bit-width is required in the PoT W[$b^\prime$]A[$b$] multiplication realized by left shifting the input activation by $b^\prime$ bits.

\subsection{Compute Engine for Multi-Head Attention}\label{sec:compute_engine}

\begin{figure}[htb]
  \centering
  \includegraphics[width=0.9\linewidth]{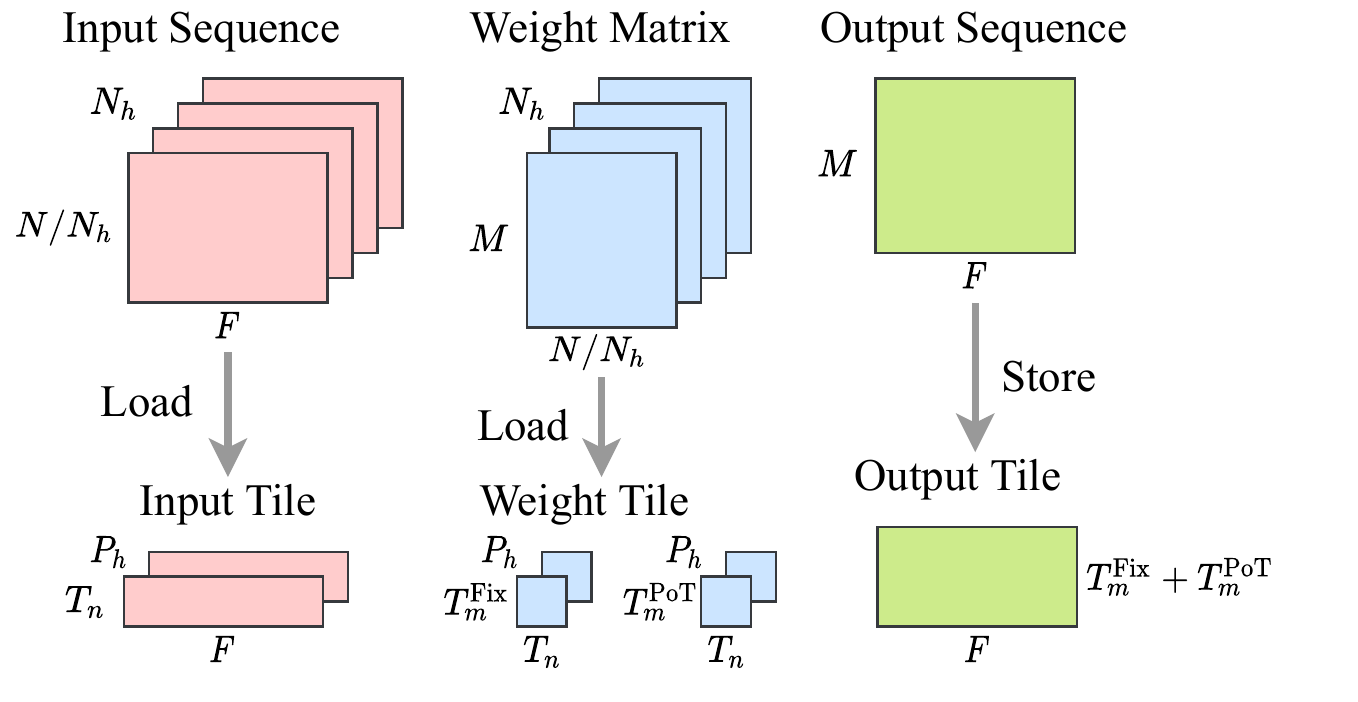}
  \caption{Tiling in ViT computations.}
  \label{fig:tiling}
\end{figure}

The notations used in ViT accelerators are listed in Table~\ref{tab:notation}.
The accelerator designs are based on loop tiling shown in Fig.~\ref{fig:tiling}, where the input, weight, and output data for each ViT layer are split into tiles for FPGA resource-saving. With pipelining and unrolling of loops, the compute engine can manage $(T_m^{\mathrm{Fix}} + T_m^{\mathrm{PoT}}) \cdot P_h \cdot T_n$ multiply-accumulate (MAC) operations in parallel.

\begin{table}[htb]
\tabcolsep 3pt
\centering

\caption{Notations for ViT Accelerator}
\label{tab:notation}

\scalebox{0.95}
{
\begin{tabular}{l|l}
\toprule
Notation & Description \\
\midrule
$k_{\mathrm{PoT}}$ & The ratio of PoT quantized rows for weights \\ \hline

$M$ ($N$) & Number of output (input) channels \\ \hline
$F$ & Number of token sequences \\ \hline

$T_n$ & Tiling size for data in input channel dimension in each head \\ \hline
$T_m^{\mathrm{Fix}}$ ($T_m^{\mathrm{PoT}}$) & Tiling size for Fixed (PoT) data in output channel dimension \\ \hline
$N_h$ & Total number of heads \\ \hline
$P_h$ & Number of heads for computation in parallel \\ \hline
\multirow{2}{*}{$D$ ($D^\prime$)} & Number of data packed as one for activations and \\ 
~ & Fixed weights (PoT weights) \\ \hline

$A_{\mathrm{in}}$ & Number of AXI ports used for data transfer of input \\
($A_{\mathrm{out}}$, $A_{\mathrm{wgt}}$) & (output, weight) tile \\ \hline
$L_{\mathrm{in}}$ ($L_{\mathrm{wgt}}$, & Number of clock cycles for input transfer (weight transfer, \\
$L_{\mathrm{out}}$, $L_{\mathrm{cmpt}}$) & output transfer, computation) for a group of tiles \\ \hline
$B_{\mathrm{in}}$ & \multirow{2}{*}{Number of BRAMs used by input (output, weight) tile} \\
($B_{\mathrm{out}}$, $B_{\mathrm{wgt}}$) & ~ \\ \hline
$C_{\mathrm{dsp}}^{\mathrm{Fix}}$ & DSP cost for each MAC operation with Fixed weight \\
$C_{\mathrm{lut}}^{\mathrm{Fix}}$ ($C_{\mathrm{lut}}^{\mathrm{PoT}}$) & LUT cost for each MAC operation with Fixed (PoT) weight \\

\bottomrule
\end{tabular}
}
\end{table}

ViT computations mainly comprise matrix multiplications in multi-layer perceptron (MLP) modules and multi-head self-attention (MSA) modules. 
Each MSA can be seen as multiple parallel matrix multiplications, and therefore the accelerator is designed to process $P_h$ attention heads in parallel, by splitting the $N$ input channels into $N_h$ groups.
This input channel splitting is also done for fully connected (FC) layers, each containing only one matrix multiplication for compatibility, and the results need to be accumulated from all the input channels in all the heads.

\subsection{Optimizations in ViT Accelerator} \label{sec:other_layers}

\subsubsection{Processing of Other Computations}

In addition to matrix multiplications, ViTs contain convolution, scaling, softmax, activation, normalization (LN), and skip-connection addition operations.
The first layer of a ViT is a convolutional layer that can be converted to an FC layer because its kernel size and stride are the same as the patch size, meaning that the input data are used only once when a weight kernel slides across the input feature map.
The scaling, softmax, and GELU activation operations are performed on the host CPU of the FPGA, which introduces a small latency overhead for embedded FPGAs compared with matrix multiplications.

As illustrated in Fig.~\ref{fig:encoder}, LN is applied at the beginning of each MSA or MLP module.
The LN inputs require to be stored for later additions due to the identity skip-connection linking the input activations of each LN and the output activations of the subsequent module.
Considering that keeping LN operations unquantized will not incur much computation overhead but help maintain the model accuracy, the LN parameters and inputs are represented with 16-bit precision on hardware.
Two data transfer ports are needed respectively for unquantized LN input and quantized LN outputs (which are also inputs of the next FC layer) to minimize the input loading time for subsequent FC computations.

\subsubsection{DSP Packing}

To fully exploit the potential of DSP resources on FPGAs, we pack multiple low-bit multiplications within each DSP following~\cite{Xilinx-dspPack-Int8,Xilinx-dspPack-Int4}.
Each DSP (DSP48E2) on the ZCU102 board could support the computation of $P=(A+D)\times B$, where both $A$ and $D$ are 27-bit operands, $B$ is an 18-bit operand, and $P$ is the 45-bit output.
One DSP can accommodate two $8 \times 8$-bit multiplications by holding one weight in A and two input activation values in B, or four $4 \times 4$-bit multiplications by holding one weight in A, another weight in D, and two inputs in B.
It is worth noting that the number of $4 \times 4$-bit multiplications handled by each DSP in this design is higher than that in~\cite{chang2021mix}, resulting in higher resource utilization efficiency and throughput.

\subsection{ViT Accelerator Design with Resource Modeling and Performance Analysis} \label{sec:accl_framework}

An FPGA board contains primarily two types of computation resources, namely DSPs and LUTs. Multiplications with fixed-point weights are computed with DSPs, and those with PoT weights can be replaced by shifting operations that are computed with LUTs. The DSP and LUT cost requires to be precisely estimated to find the best ratio between the numbers of fixed-point and PoT weights and thus maximizing the throughput on FPGAs.

The parameters to be determined for the accelerator include $T_m^{\mathrm{Fix}}$ ($T_m^{\mathrm{PoT}}$), $T_n$, $D$ ($D^\prime$), and $P_h$.
On a specific FPGA board, the maximum achievable FPS, denoted by $\mathrm{FPS_{max}}$, can be estimated according to our analysis of FPGA resource utilization and performance.
Given the target FPS, denoted by $\mathrm{FPS_{tgt}}$, we first find the precision and scheme combination satisfying $\mathrm{FPS_{max}} \geq \mathrm{FPS_{tgt}}$.
Under this precision, we fix $P_h$, $T_n$, $D$ ($D^\prime$), and $T_m^{\mathrm{Fix}}$, and adjust $T_m^{\mathrm{PoT}}$ to meet the target FPS and obtain the best model accuracy.
In detail, $P_h$ is set to a value that can divide $N_h$ exactly for full exploitation of computation resources, i.e., $P_h=3$ for $N_h=6$, and $P_h=4$ for $N_h=8$ or $N_h=12$.
$D$ is decided based on the FPGA AXI port size and the quantization bit-width of Fixed weights, and is the same for activations in both Fixed and PoT computations as well as weights in Fixed computations. The bit-width of PoT weights is lower, corresponding to $D^\prime$. $T_n$ is set to the same value as $D$.
The computation parallelism along the output channel dimension is decided by the sum $T_m^{\mathrm{Fix}}+T_m^{\mathrm{PoT}}$, and the model accuracy in quantization is affected by the ratio $k_{\mathrm{PoT}} = \frac{T_m^{\mathrm{PoT}}}{T_m^{\mathrm{PoT}} + T_m^{\mathrm{Fix}}}$, i.e., lower $k_{\mathrm{PoT}}$ will result in higher model accuracy.
We therefore reduce $T_m^{\mathrm{PoT}}$ to make the actual FPS equal to $\mathrm{FPS_{tgt}}$ if $\mathrm{FPS_{max}} > \mathrm{FPS_{tgt}}$ under this precision, and the actual $k_{\mathrm{PoT}}$ ratio will guide the quantization process and the hardware implementations with all these parameters.

\subsubsection{FPGA Resource Utilization Modeling}

In contrast to DSP usage, LUT consumption for shifting operations and also for logic is difficult to estimate, and therefore we build a resource utilization model through several simple experiments to model the LUT cost as a linear function of computation parallelism (the number of parallel operations in each clock cycle).
For Fixed W[$b$]A[$b$] + PoT W[$b^\prime$]A[$b$] quantization, the LUT cost is analyzed for both W[$b$]A[$b$] Fixed multiplications executed on DSPs (denoted by $C_{\mathrm{lut}}^{\mathrm{Fix}}$), and W[$b^\prime$]A[$b$] PoT multiplications executed on LUTs (denoted by $C_{\mathrm{lut}}^{\mathrm{PoT}}$). The LUT cost can then be obtained from the slopes of the fitted lines.
It is worth noting that employing DSPs for multiplications consumes LUTs as well, resulting from data packing and accumulation operations, etc.

\subsubsection{Inference Latency Analysis} \label{sec:opt_latency}

\begin{table*}[htb]
\centering
\tabcolsep 3pt

\caption{Accuracy and Hardware Results under Different Quantization Schemes for DeiT-small and DeiT-base on ImageNet Dataset}\label{tab:comparison_schemes}

\scalebox{0.95}
{
\begin{tabular}{cc|cc|cc|cccc}
\toprule
Quantization & Bit-Width & \multicolumn{2}{c|}{Model Accuracy (\%)} & \multicolumn{2}{c|}{Resource Utilization} & Power & Thrpt. & Frame Rate & Energy Eff. \\
Weight Scheme & (Weight/Actv.) & Top-1 & Top-5 & DSP & kLUT & (W) & (GOPS) & (FPS) & (FPS/W) \\
\hline
\multicolumn{10}{c}{\textbf{DeiT-small}} \\
\hline
Baseline & W32A32 &  79.85&  94.97 & 1745 (69\%) & 130 (47\%) & 8.38 & 354.5 & 38.9 & 4.64 \\
PTQ~\cite{liu2021post} (Fixed) & W8A8 & 77.47 ($-$2.38) & - & - & - & -& - & - & -  \\
Fixed & W4A4 & 78.50 ($-$1.35)&  94.41 ($-$0.56) & 1933 (77\%) & 137 (50\%) & 10.44 & 1186.6 & 130.3 & 12.48 \\
\bf{PoT} & W3A4 & 77.24 ($-$2.61) & 93.89 ($-$1.08) & 13 (1\%) & 176 (64\%) & 6.55 & 1374.1 & 150.9 & 23.04 \\
\bf{Mixed ($\mathrm{FPS_{tgt}}=150$)} & W4A4$+$W3A4 ($k_{\mathrm{PoT}}=43\%$) & 77.94 ($-$1.91) & 94.07 ($-$0.90) & 1549 (61\%) & 193 (70\%) & 10.34 & 1418.4 & 155.8 & 15.06 \\
Fixed & W8A8 &  79.69 ($-$0.16) & 94.89 ($-$0.08) & 1936 (77\%) & 122 (44\%) & 8.46 & 711.2 & 78.1 & 9.23 \\
\bf{PoT} & W4A8 & 77.97 ($-$1.88) & 94.06 ($-$0.91) & 16 (1\%) & 175 (64\%) & 8.58 & 837.0 & 91.9 & 10.71 \\
\bf{Mixed ($\mathrm{FPS_{tgt}}=100$)} & W8A8$+$W4A8 ($k_{\mathrm{PoT}}=43\%$) &  78.74 ($-$1.11)&  94.50 ($-$0.47) & 1552 (62\%) & 185 (67\%) & 9.63 & 907.8 & 99.7 & 10.35 \\
\hline
\multicolumn{10}{c}{\textbf{DeiT-base}} \\
\hline
Baseline & W32A32 & 81.85&  95.59 & 1564 (62\%) & 120 (44\%) & 9.91 & 345.8 & 10.0 & 1.01 \\
PTQ~\cite{liu2021post} (Fixed) & W8A8 & 80.48 ($-$1.37) & - & - & - & -& -& -& - \\
Fixed & W4A4 & 81.33 ($-$0.52)&  95.63 ($+$0.06) & 2064 (82\%) & 139 (51\%) & 11.27 & 1648.1 & 47.5 & 4.21 \\
\bf{PoT} & W3A4 & 80.87 ($-$0.98) & 95.57 ($-$0.02) & 19 (1\%) & 191 (70\%) & 8.11 & 1958.4 & 56.4 & 6.95 \\
\bf{Mixed ($\mathrm{FPS_{tgt}}=50$)} & W4A4$+$W3A4 ($k_{\mathrm{PoT}}=40\%$) & 81.14 ($-$0.71) & 95.60 ($+$0.01) & 1555 (62\%) & 179 (65\%) & 11.03 & 1970.3 & 56.8 & 5.15 \\
Fixed & W8A8 & 81.93 ($+$0.08) & 95.90 ($+$0.31) & 2066 (82\%) & 128 (47\%) & 9.40 & 899.6 & 25.9 & 2.76 \\
\bf{PoT} & W4A8 & 81.51 ($-$0.34)& 95.73 ($+$0.14) & 20 (1\%) & 192 (70\%) & 7.24 & 1080.5 & 31.1 & 4.30 \\
\bf{Mixed ($\mathrm{FPS_{tgt}}=30$)} & W8A8$+$W4A8 ($k_{\mathrm{PoT}}=45\%$) & 81.84 ($-$0.01)&  95.85 ($+$0.26) & 1556 (62\%) & 186 (68\%) & 9.31 & 1181.5 & 34.0 & 3.66 \\

\bottomrule
\end{tabular}
}
  \vspace{-0.4cm}
\end{table*}

The actual FPS is the reciprocal of the inference latency, which is analyzed below, with main variables explained in Table~\ref{tab:notation}.
For one layer $i$ in ViTs, the numbers of clock cycles needed for input tile loading, weight tile loading, and output tile storage are calculated as
\begin{equation}
\small
\begin{aligned}
L_{\mathrm{in}} &= P_h \cdot \left\lceil \frac{T_n}{D} \right\rceil \cdot \left\lceil \frac{F}{A_{\mathrm{in}}} \right\rceil, \\
L_{\mathrm{wgt}} &= P_h \cdot \bigg( \left\lceil \frac{T_n}{D} \right\rceil \cdot \left\lceil \frac{T_m^{\mathrm{Fix}}}{A_{\mathrm{wgt}}} \right\rceil + \left\lceil \frac{T_n}{D^\prime} \right\rceil \cdot \left\lceil \frac{T_m^{\mathrm{PoT}}}{A_{\mathrm{wgt}}} \right\rceil \bigg), \\
L_{\mathrm{out}} &= (1 + \gamma) \cdot \left\lceil \frac{T_m^{\mathrm{Fix}} + T_m^{\mathrm{PoT}}}{D} \right\rceil \cdot \left\lceil \frac{F}{A_{\mathrm{out}}} \right\rceil,
\end{aligned}
\end{equation}
where $\gamma$ is $N_h-1$ if the current layer is a multi-head attention layer else 0.
Additionally, the clock cycle number of computations for one group of tiles is
\begin{equation}
\small
L_{\mathrm{cmpt}} = \left\lceil \frac{F}{2} \right\rceil \cdot \left\lceil \frac{N_h}{P_h} \right\rceil,
\end{equation}
as two input values are fetched in each clock cycle for DSP packing.
The data loading and computation for the tiles are conducted simultaneously with the double buffering technique to overlap the data transfer with computations. The clock cycle number of this process is $L_1 = \max\{L_{\mathrm{in}}, L_{\mathrm{wgt}}, L_{\mathrm{cmpt}}\}$.
And to obtain the accumulation of output results, this process is performed multiple times. The clock cycle number for calculating the whole output tile is $L_2 = \max \big\{L_1 \cdot \left\lceil \frac{N}{P_h \cdot T_n} \right\rceil + L_{\mathrm{cmpt}}, L_{\mathrm{out}} \big\}$.
The total number of clock cycles for a ViT layer $i$ is therefore described by
\begin{equation}
\small
L_{\mathrm{tot}}^i = \left\lceil \frac{M}{T_m^{\mathrm{Fix}} + T_m^{\mathrm{PoT}}} \right\rceil \cdot L_2 + L_{\mathrm{out}}.
\label{clock_cycle}
\end{equation}
Under a working frequency $f$, the $\mathrm{FPS}$ is calculated as $\frac{f}{\sum\limits_i L_{\mathrm{tot}}^i}$.

With double buffering, the 18k-bit BRAM usage of the input, weight, and output tiles are given by
\begin{equation}
\small
\scalebox{0.8}{$
\begin{aligned}
B_{\mathrm{in}} &= 2 \cdot P_h \cdot \left\lceil \frac{T_n}{D} \right\rceil \cdot \left\lceil \frac{b \cdot F \cdot D}{18\mathrm{k}} \right\rceil, \\
B_{\mathrm{wgt}} &= 2 \cdot P_h \cdot \bigg( \left\lceil \frac{T_n}{D} \right\rceil \cdot \left\lceil \frac{b \cdot T_m^{\mathrm{Fix}} \cdot D}{18\mathrm{k}} \right\rceil + \left\lceil \frac{T_n}{D^\prime} \right\rceil \cdot \left\lceil \frac{b^\prime \cdot T_m^{\mathrm{PoT}} \cdot D^\prime}{18\mathrm{k}} \right\rceil \bigg), \\
B_{\mathrm{out}} &= 2 \cdot N_h \cdot \left\lceil \frac{T_m^{\mathrm{Fix}} + T_m^{\mathrm{PoT}}}{D} \right\rceil \cdot \left\lceil \frac{b \cdot F \cdot D}{18\mathrm{k}} \right\rceil.
\end{aligned}
$}
\end{equation}
The DSP and LUT consumption is proportional to the total MAC computation parallelism. Specifically, $C_{\mathrm{dsp}}^{\mathrm{Fix}}=0.25$ for each multiplication with W4A4 or smaller precision, and $C_{\mathrm{dsp}}^{\mathrm{Fix}}=0.5$ for each multiplication with W5A5 to W8A8 precision.
In summary, the FPS and $k_{\mathrm{PoT}}$ for the ViT are decided satisfying
\begin{equation}
\small
\begin{aligned}
B_{\mathrm{in}} + B_{\mathrm{wgt}} + B_{\mathrm{out}} &\leq S_{\mathrm{bram}}, \\
C_{\mathrm{\mathrm{dsp}}}^{\mathrm{Fix}} \cdot T_m^{\mathrm{Fix}} \cdot P_h \cdot T_n &\leq S_{\mathrm{dsp}} \cdot r_{\mathrm{dsp}}, \\
\big( C_{\mathrm{lut}}^{\mathrm{Fix}} \cdot T_m^{\mathrm{Fix}} + C_{\mathrm{lut}}^{\mathrm{PoT}} \cdot T_m^{\mathrm{PoT}} \big) \cdot P_h \cdot T_n &\leq S_{\mathrm{lut}} \cdot r_{\mathrm{lut}},
\end{aligned}
\label{equ:constraint}
\end{equation}
where $S_\mathrm{bram}$, $S_\mathrm{dsp}$, $S_\mathrm{lut}$ are the available number of BRAMs, DSPs, and LUTs on FPGA, and $r_{\mathrm{dsp}}$ and $r_{\mathrm{lut}}$ are the maximum ratio of DSPs and LUTs to be utilized for MAC operations.

\section{Experiments}
\subsection{Experimental Setups}
Our experiments include model quantization and hardware implementations for ViTs of different sizes, namely DeiT-small and DeiT-base, without the distillation tokens~\cite{Touvron2021TrainingDI}.
Our quantization training process takes 100 epochs with a batch size of 512, on top of the pre-training process with 300 epochs. The learning rate is set to $5 \times 10^{-4}$ initially and decayed with a cosine annealing schedule. The AdamW~\cite{loshchilov2018decoupled} optimizer is used with the weight decay of 0.05. Training tricks to improve the accuracy include warmup training of 3 epochs and label smoothing with a factor of 0.1. 
The quantization adopts the same hyper-parameters for all models and is conducted on 4 NVIDIA Ampere A100 GPUs with CUDA 11.0 and PyTorch 1.7 frameworks on the Ubuntu operating system.
The quantized models are then evaluated on the Xilinx ZCU102 FPGA platform consisting of 2520 DSPs and 274.1k LUTs. To maximize the computation efficiency without timing violation, the working frequency is set to 150 MHz for all the designs implemented through Xilinx Vitis and Vitis HLS 2020.2.
\textcolor{blue}{
We use the official DeiT model (W32A32) as our baseline. And the W32A32 data in baseline unquantized models are represented in 16-bit format when implemented on FPGA. This conversion incurs negligible accuracy degradation, which is common for FPGA implementations.
}

\subsection{Experimental Results}
\noindent{\bf Comparison of Different Quantization Schemes.}
The comparison results of different quantization schemes in terms of accuracy after quantization and performance with resource utilization are listed in Table~\ref{tab:comparison_schemes}. All the activation are quantized with Fixed schemes, and the weights are quantized with the schemes as shown in the first column.
It can be seen that the PoT quantization on ViTs obtains noticeable throughput improvement compared with the Fixed-point quantization at the same bit-width level with manageable accuracy loss, and our mixed-scheme quantization further achieves higher throughput and better model accuracy than the PoT quantization. With various FPS targets, we investigate the effectiveness of our mixed-scheme quantization by adjusting the bit-widths and scheme mixing ratio for different models. Specifically, we set the target FPS as 150 and 100 for DeiT-small, and 50 and 30 for DeiT-base.

For DeiT-small, it can be seen that a target FPS of 150 can be met using W4A4+W3A4 quantization precision with PoT ratio $k_{\mathrm{PoT}}=43\%$ and the Top-1 accuracy reaches 77.94\%, outperforming the W8A8 model of PTQ~\cite{liu2021post} by 0.47\% even with a lower bit-width. For the desired FPS of 100, the implementation using W8A8+W4A8 precision with PoT ratio of $k_{\mathrm{PoT}}=43\%$ can fulfill the requirement with 78.74\% accuracy, which is 1.27\% higher than that of PTQ.
As for DeiT-base, the accuracy loss incurred by quantization is less than 1\%, while 55 FPS with 81.14\% accuracy can be achieved using W4A4+W3A4 precision with $k_{\mathrm{PoT}}=40\%$, and 33 FPS with 81.84\% accuracy can be reached using W8A8+W4A8 precision with $k_{\mathrm{PoT}}=45\%$.

\noindent{\bf Comparison with Baseline and Other Framework.}
Under the similar quantization bit-width, the Top-1 accuracy of our Fixed W8A8 + PoT W4A8 model is 1.36\% higher than that in PTQ. Under a lower bit-width, our Fixed W4A4 + PoT W3A4 model still outperforms the W8A8 model of PTQ by 0.66\%.
Compared with the 32-bit baseline model, our quantized model achieves around 5.6$\times$ 
improvement on frame rate (i.e., 56.8 FPS vs. 
10.0 FPS) with only 0.71\% Top-1 accuracy drop.

\begin{table}[h]
\centering
\caption{
\textcolor{blue}{
Performance comparison between TX2 and ZCU102 (FPGA) on full-precision models.
}
}
\begin{tabular}{ccccc}
\toprule
\multicolumn{1}{c|}{Model} & Hardware & Power (W) & Latency (ms) & FPS \\
\midrule
\multicolumn{1}{c|}{\multirow{2}{*}{DeiT-small}} & TX2 & 11.87 & 54 & 18.52 \\
\multicolumn{1}{c|}{}  & ZCU102 & 8.38 & 26 & 38.90 \\
\midrule
\multicolumn{1}{c|}{\multirow{2}{*}{DeiT-base}} &  TX2  & 12.28 & 127 & 7.87 \\
\multicolumn{1}{c|}{} &  ZCU102 & 9.91 & 100 & 10.00 \\
\bottomrule
\end{tabular}
\label{TX2-vs-FPGA}
\end{table}

\noindent{\bf Comparison with Embeded CPU/GPU.}
\textcolor{blue}{
We also test DeiT-base and DeiT-small on Jetson TX2 with 4-core ARM CPU and NVIDIA Pascal GPU, and compared them with our FPGA (ZCU102) implementation. Since TX2 GPU does not support low-bit computation, we only present the performance of the full precision model as shown in Table~\ref{TX2-vs-FPGA}. Overall, compared to TX2 GPU, our FPGA implementation achieves about 2x and 1.3x speedup on DeiT-small and DeiT-base, respectively, with 2.37 W~3.49 W lower power consumption. Even without quantization, our FPGA implementation is still more efficient compared with TX2 with the similar compute capability level.
}


\section{Conclusion}
\textcolor{blue}{
In this paper, we propose an FPGA-aware automatic ViT acceleration (\M) framework for our mixed-scheme ViT quantization algorithm. The bit-width and the ratio of fixed-point quantized rows over PoT quantized rows can be optimized and used as inputs to guide the quantization algorithm.
This framework also designs a novel FPGA compute engine for ViT multi-head attention with optimizations for accelerators.
We automate the entire workflow based on a target FPS, to obtain a quantized model and an FPGA accelerator.
Compared with the 32-bit floating-point baseline FPGA accelerator, our accelerator achieves around 5.6$\times$ improvement on the frame rate with 0.71\% accuracy drop on ImageNet dataset for DeiT-base.
}
To the best of our knowledge, this is the first work for quantization-based ViT acceleration on FPGAs.

\section*{Acknowledgment}
\textcolor{blue}{
This work is partly supported by NSF CCF-1901378;
}
\textcolor{red}{
NSERC Discovery Grant RGPIN-2019-04613, DGECR-2019-00120, Alliance Grant ALLRP-552042-2020; CFI John R. Evans Leaders Fund.
}

\bibliographystyle{IEEEtran}
\bibliography{ref}

\end{document}